\documentclass{article}

\usepackage[pdftex]{graphicx}
\usepackage{amsmath,amssymb} 

\DeclareMathOperator{\sign}{sign}

\usepackage[final]{nips_2018}

\usepackage[utf8]{inputenc} 
\usepackage[T1]{fontenc}    
\usepackage{hyperref}       
\usepackage{url}            
\usepackage{booktabs}       
\usepackage{amsfonts}       
\usepackage{nicefrac}       
\usepackage{microtype}      

\title{Rank Projection Trees for Multilevel Neural\\ Network Interpretation}

\author{
 Jonathan Warrell$^{1,2}$, Hussein Mohsen$^{1,2}$, and Mark Gerstein$^{1,2,3}$\\
 $^1$ Program in Computational Biology and Bioinformatics,  Yale University\\
 $^2$ Molecular Biophysics and Biochemistry,  Yale University\\
 $^3$ Department of Computer Science,  Yale University\\
 \texttt{jonathan.warrell@\{gmail.com, yale.edu\},} \texttt{hussein.mohsen@yale.edu,}\\ 
 \texttt{mark@gersteinlab.org}\\
}

\begin{document}

\maketitle

\begin{abstract}
A variety of methods have been proposed for interpreting nodes in deep neural networks, which typically involve scoring nodes at lower layers with respect to their effects on the output of higher-layer nodes (where lower and higher layers are closer to the input and output layers, respectively).  However, we may be interested in picking out a prioritized collection of subsets of the inputs across a range of scales according to their importance for an output node, and not simply a prioritized ranking across the inputs as singletons.  Such a situation may arise in biological applications, for instance, where we are interested in epistatic effects between groups of genes in determining a trait of interest.  Here, we outline a flexible framework which may be used to generate multiscale network interpretations, using any previously defined scoring function. We demonstrate the ability of our method to pick out biologically important genes and gene sets in the domains of cancer and psychiatric genomics.

\end{abstract}

\section{Introduction}

Interpretation of deep neural networks is an important problem in domains such as computational biology, where understanding the features that a network is using to predict a disease for instance may shed light on underlying biological mechanisms.  Numerous schemes have been proposed for network interpretation, which typically involve scoring nodes at lower layers with respect to their effects on the output of higher-layer nodes. Methods which have been proposed include gradient-based schemes [1,2], gradient+input schemes [3], perturbation schemes [4,5], and difference scores with respect to a reference [6]. Each of these schemes has benefits and drawbacks, including for instance  computational efficiency, ability to cope with saturated inputs and non-differentiability, and the need for a reference. In addition, such methods all involve scoring individual nodes at a layer of interest, while we may be interested in picking out a collection of subsets of nodes at multiple scales from a given layer which are `important' to the network.

Such a situation often arises in biological applications where we are interested in epistatic effects between groups of genes in determining a trait of interest.  For instance, the framework of {\em weighted gene coexpression network analysis} (WGCNA, see [7,8]) is widely used to group together genes sharing common coexpression patterns, whose relevance to particular high-level traits (for instance psychiatric conditions [9]) may be determined. The `modules' found by WGCNA however are constrained to be non-overlapping, meaning that the same gene cannot participate in groupings at multiple levels. Additionally, WGCNA does not use trait-relevant information in defining modules.

Here, we outline a flexible framework ({\em rank projection trees}) which can be used to select collections of trait-relevant subsets of interacting genes by providing a multiscale interpretation of a supervised deep neural network. Our framework is agnostic as to the underlying ranking function which is used to build the tree, allowing any of the above methods to be used to provide a scoring function.  We demonstrate the ability of our method to pick out biologically important genes and gene sets in the domains of cancer and psychiatric genomics, using networks learned on genomics data from PCAWG [10] and PsychENCODE [11,12] consortia to predict epistatic interactions of germline and somatic mutations in cancer, and risk for Schizophrenia, respectively.

\section{Rank Projection Trees}

\begin{figure*}[!t]
\centering
\includegraphics[width=3.9in]{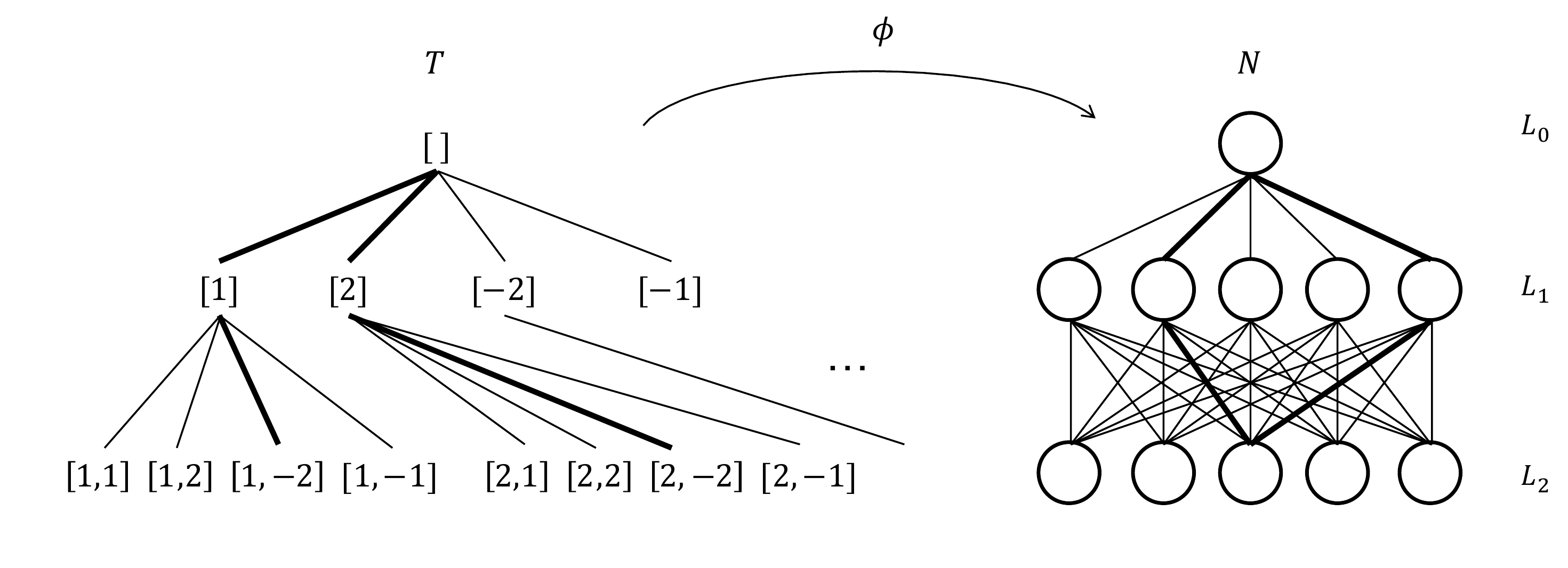}
\caption{Rank projection tree schematic.  The rank projection tree (left, $T$) is mapped onto a trained neural network (right, $N$) via the mapping $\phi$ which depends on an arbitrary ranking function $r$.  The image of $T$ under $\phi$ is used to prioritize inputs and sets of inputs in $N$ in an output dependent fashion.}
\label{fig:fig1}
\end{figure*}

We assume we have a neural network $N$ with layers $l=0...L$, where layers $0$ and $L$ are the output and input layers, respectively. Each layer $l$ has an associated set $I_l = \{1...N_l\}$ indexing the nodes of that layer; hence $n_{l,i}$ is the $i$'th node at layer $l$.  For convenience, we assume there is one output node, $n_{0,1}$.  We write the weight matrix between layers $l_1$ and $l_2=l_1-1$ as $W_{l_1,l_2}$, and the biases vector at layer $l$ as $\beta_l$.  A {\em rank projection tree} (see Fig. \ref{fig:fig1}) over a given network is fully determined by specifying (a) a half branching factor $B<N_l/2, \forall l$, and (b) a ranking function $r_{i,l,m}(j)$, where $l<m\leq L$ are layer indices, $i$ and $j$ are node indices on layers $l$ and $m$ respectively, and the function returns an integer specifying the position of node $j$ in an ordering of the nodes at layer $m$ according to their `score' with respect to node $i$ and layer $l$.  Semantically, we expect that increased activation of node $j$ towards the top of the ranking will lead to increased activation of node $i$, while increased activation of nodes towards the bottom will lead to decreased activation of $i$; hence, any score function of the kind described above (such as the gradient) may be used.

The nodes of the rank projection tree $T$ are lists of `branching indices' of the form $[\;]$, $[b_1]$, $[b_1,b_2]$, ... $[b_1,...,b_L]$, where $b_l \in \{1,...,B\}\cup\{-1,...,-B\}, \; \forall l$.  The node $[\;]$ is the root of the tree, and the parent function is defined as $Pa([b_1,...,b_{l-1},b_l]) = [b_1,...,b_{l-1}]$.  A node $t$ of $T$, where $t$ is a list of length $l$, is then associated with a node in layer $l$ of the neural network via a function $\phi$ defined recursively as follows:
\begin{eqnarray}\label{eq:phi}
&&\phi : T \rightarrow N \nonumber \\
&&\phi([\;]) = n_{0,1}  \nonumber \\
&&\phi(t=[b_1,...b_l]) = r^{-1}_{Pa(t),l-1,l}(b_l),
\end{eqnarray}
\noindent where $r^{-1}_{i,l,m}(b)$ is a `quasi-inverse' of the ranking function, which returns the node $n_{m,j}$ for which $r_{i,l,m}(j)=b$ if $b>0$, and $n_{m,j}$ for which $r_{i,l,m}(j)=N_m+b+1$ if $b<0$.  For node $t$ in $T$ then, which maps to $\phi(t)=n$ at layer $l$, the mappings of the children of $t$ are set by first ranking layer $l+1$ of the neural net with respect to $n$, and assigning the top $B$ and bottom $B$ nodes of this ranking to the children of $t$, hence {\em projecting} the full ranking onto a reduced ranking across the $2B$ children.  

Each node $t$ in $T$ may be associated with positive and negative subsets, $S^+_t$ and $S^-_t$, at a reference layer, which we take to be the input layer $L$.  These are defined as:
\begin{eqnarray}\label{eq:posnegsets}
S^+_t &=& \{\phi(t'=[t,b_{l+1},b_{l+2},...,b_L])| b_{l+1}\cdot b_{l+2} ...\cdot b_L > 0 \} \nonumber\\
S^-_t &=& \{\phi(t'=[t,b_{l+1},b_{l+2},...,b_L])| b_{l+1}\cdot b_{l+2} ...\cdot b_L < 0 \}.
\end{eqnarray}
Hence, $S^+_t$ contains all those nodes mapped to by descendants of $t$ at layer $L$ along paths where the product of the branching indices below $l$ is positive, and $S^-_t$ is defined similarly, but where the product of the branching indices is negative. A collection of `prioritized' subsets at multiple levels is thus formed by applying Eq. \ref{eq:posnegsets} as $t$ runs across $T$.  We note that, since multiple nodes in $T$ may map to the same node in $N$, sets at the same layer may overlap, including positive and negative sets associated with the same node in $T$.  Finally, we may define a prioritization function $\pi$ (or `salience map') of the nodes at the reference layer in $N$, $\pi(n) = f(\phi^{-1}(n))$, where $\phi^{-1}(n)=\{t_1,t_2...\}$ is the pre-image of $n$ under $\phi$, and $f$ may be chosen from a number of possibilities we explore below.  

\section{Results}

In the following, we explore the ability of the rank projection tree defined above to prioritize genes and subsets of genes relevant in cancer and psychiatric genomics applications.  For cancer genomics, we use data from the PanCancer Analysis of Whole Genomes dataset (PCAWG, [10]) to train neural networks (3 hidden layers) to predict the per-tumor co-occurrence of germline and somatic mutations in a given gene (a binary output), using germline variant signatures of known cancer genes alongside a set of gene-level biological features as inputs.  For psychiatric genomics, we use data from the PsychENCODE dataset [11,12] to train neural networks (2 hidden layers) to classify control versus Schizophrenia post-mortem subjects after balancing covariates (age, gender, ethnicity, assay), using bulk transcriptomic data from the prefrontal cortex as inputs, either in the form of individual gene expression levels, or average expression across predefined modules of genes (using WGCNA [7,8]).  Further details on the datasets and training of the networks are in Appendix A.

\subsection{Prioritization functions}

\begin{table}
  \caption{Prioritization function comparison.  The rankings of genes induced by different prioritization functions are compared against citation-based rankings from existing literature.  Table shows normalized $\ell_1$-distances of predicted and citation-based rankings for the top 20 genes, with best performing metrics highlighted.  Rows are: Schizophrenia, Skin Melanoma and Esophageal Adenocarcinoma.}
  \label{table1}
  \centering
  \begin{tabular}{cccccccccc}
    \toprule
     Network & \multicolumn{4}{c}{Absolute} & \multicolumn{4}{c}{Signed} & \multicolumn{1}{c}{Count}\\ 
     \cmidrule(l){2-5} \cmidrule(l){6-9}
        &  Sum & Average & Max & Min  &  Sum & Average & Max & Min  &  \\
    \midrule
    Schiz.  & 0.642 & \textbf{0.628} & 0.664 & 0.656 & 0.710 & 0.674 & 0.708 & 0.686 & 0.634   \\
    Skin Mel.     & 0.628 & \textbf{0.583} & 0.632 & 0.624 & 0.588 & 0.597 & 0.614 & 0.625 & 0.614      \\
    Es. Adeno.  & 0.655 & 0.644 & \textbf{0.603} & 0.661 & 0.670 & 0.612 & 0.642 & 0.671 & 0.670     \\
    \bottomrule
  \end{tabular}
\end{table}

We first compare the ability of different prioritization functions to pick out individual genes relevant to the diseases predicted by each network.  For this purpose, we use a simple ranking function where, for layers $l$ and $m=l+1$, $r_{i,l,m}(j)$ returns the rank of $n_{m,j}$ in the ordering induced by the weights $W_{lm}(i,.)$ (signed values, descending), and set $B=2$. For the prioritization function $\pi$, we first calculate the cumulative rank-scores $c_t$ of each path in $T$; that is, for leaf node $t = [b_1,...b_L]$, we set $c_t=(\sum_l |b_l|)(\prod_l \sign(b_l))$.  We then set $\pi(n) = f(\phi^{-1}(n)) = g([c_{t_1},c_{t_2},...])$ for $t_1,t_2... \in \phi^{-1}(n)$, where $g(.)$ is one of the functions: \{sum, average, max, min\} of signed or absolute values, or length.  Table \ref{table1} scores the rankings of the top 20 genes (averaged across networks) according to each $g(.)$ when compared against a `ground truth ranking'  based on the number of citations retrieved by Google Scholar when each gene is queried in association with the disease; the table shows the $\ell_1$-distance between the two rankings, normalized by its maximum ($\ell_1([1...20],[20...1])$).  In general, the absolute average and max functions appear to be better indicators of individual gene importance.

\vspace{-0.2cm}
\subsection{Interpreting gene groupings at multiple levels}

\begin{table}
  \caption{Annotation enrichment across layers.  Table compares the total citations of the top 20 KEGG terms associated with gene groupings at each layer of neural networks trained to predict healthy versus schizophrenia with two hidden layers. Rank-projection and randomized trees are compared.}
  \label{table2}
  \centering
  \begin{tabular}{ccccc}
    \toprule
    Model   &  $L_0$ & $L_1$ & $L_2$  &  $L_3$ \\
    \midrule
    Rank-projection & 65K & 82K & 77K & 73K    \\
    Randomized    & 53K & 54K & 56K & 72K     \\
    \bottomrule
  \end{tabular}
\end{table}

\vspace{-0.2cm}
To test the ability of the rank projection tree to extract meaningful gene groupings at multiple levels, we extracted all positive and negative groupings across layers $l=0...3$ using Eq. \ref{eq:posnegsets} for the schizophrenia networks, using exactly the same settings.  For comparison, we built identical trees, but used a randomized ranking function $r(.)$ to produce the $\phi$ mapping to the schizophrenia networks.  We used the networks with the WGCNA modules average expression levels as inputs; hence, for set $S$ formed from Eq. \ref{eq:posnegsets}, we take the union of the genes in all modules which are elements of $S$.

For our first comparison, we annotate all groupings with KEGG pathway terms using gene-set enrichment analysis [13].  All terms are assigned to a grouping achieving a q-value $<0.001$, and a ranking across KEGG terms is made for each layer independently by counting the number of groupings a term is associated with across all models (including duplicate groupings, hence accounting for increased importance of nodes in $N$ mapped to by multiple nodes in $T$).  We take the top 20 such terms, and sum the number of citations associated with these terms in association with schizophrenia by Google Scholar as above.  Table \ref{table2} compares the total citations across layers, showing that the rank projection tree finds more important terms than the random tree at all layers, and the groupings produced by higher layers of the network ($L_3$ `lowest', $L_0$ `highest') tend to associate with more trait-relevant terms (peaking at the penultimate layer, $L_1$).

\begin{figure*}[!t]
\centering
\includegraphics[width=5in]{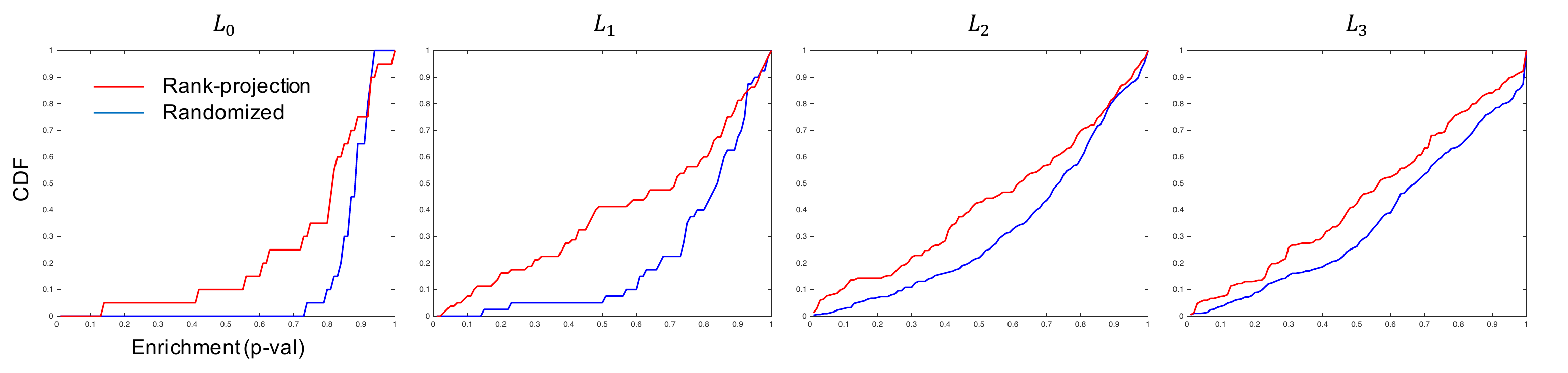}
\caption{Enrichment of Schizophrenia GWAS-linked genes across layers.  Figure compares the enrichment of high-confidence schizophrenia associated genes in gene groupings found associated with different neural network layers.  Enrichment p-values are from the hyper-geometric test, and the empirical cumulative density function (CDF) is plotted on the y-axis.}
\label{fig:fig2}
\end{figure*}

Finally, we analysed the enrichment of the groupings from all layers as described above for `high-confidence schizophrenia genes'.  These are genes which can be linked to GWAS hits for schizophrenia by any three of the following four methods: Hi-C interactions; enhancer-target links (using covariation of chromatin and expression signals); eQTL linkages, and isoform-QTL linkages (321 genes; list to be made available in [12]).  The enrichment of such genes in each module is scored using a p-value from the hyper-geometric test. Fig. \ref{fig:fig2} shows that these genes are significantly more enriched in the groupings found by the rank projection tree than randomized trees, where the groupings found by the penultimate layer $L_1$ again appear most trait-relevant, matching the findings of the citation-based metric above.  We also compared the $L_3$ distribution with a gradient-based prioritization scheme which ranked the modules at this level according to the absolute magnitude of the gradient of the network output with respect to each input (as in [1]), but found that it was not significantly better than the randomized tree (p=0.78); the rank projection tree was better than the randomized tree across all layers (p<1e-4 for $L_{1-3}$, and p=0.012 for $L_0$, all p-values using 1-tail KS-tests).

\vspace{-0.3cm}
\section{Discussion}
\vspace{-0.2cm}
We have proposed a general framework for prioritizing inputs and multiple overlapping sets of inputs for relevance to an output variable given a neural network trained for prediction. Our framework also addresses network interpretability, an issue of growing importance in the AI research community, including the representational characteristics of different network layers.  In future work, we plan to explore the relative benefits of ranking functions from the previous literature more extensively in our framework, dependence on the tree parameter $B$, and alternative forms of $f(.)$ and $S^+_t/S^-_t$ used in building the prioritization function and positive/negative groupings, respectively.  We also envisage that graph-theoretic techniques (such as analysis of the sub-graphs of the DAG image of $T$) could provide further tools for prioritizing / grouping nodes.  Finally, we plan to expand our analyses over the full range of cancer types and psychiatric conditions represented in the PsychENCODE and PCAWG datasets, which offer an ideal test-bed for the generality of methods of network interpretation.

\section*{References}

\small

[1] Simonyan, K., Vedaldi, A. and Zisserman, A. (2013) Deep inside convolutional networks: Visualising image classification models and saliency maps. arXiv preprint arXiv:1312.6034.

[2] Sundararajan, M., Taly, A., and Yan, Q. (2016) Gradients of counterfactuals. {\em CoRR}, abs/1611.02639.

[3] Shrikumar, A., Greenside, P., Shcherbina, A., and Kundaje, A. (2016) Not just a black box: Learning important features through propagating activation differences. arXiv preprint arXiv:1605.01713.

[4] Zintgraf, L. M., Cohen, T. S., Adel, T., and Welling, M. (2017) Visualizing deep neural network decisions: Prediction difference analysis. {\em ICLR}.

[5] Zhou, J. and Troyanskaya, O.G. (2015) Predicting effects of noncoding variants with deep learning–based sequence model. {\em Nature methods}, 12(10), p.931.

[6] Shrikumar, A., Greenside, P. and Kundaje, A. (2017) Learning important features through propagating activation differences. arXiv preprint arXiv:1704.02685.

[7] Zhang, B. and Horvath, S. (2005) A general framework for weighted gene co-expression network analysis. {\em Statistical applications in genetics and molecular biology}, 4(1).

[8] Langfelder, P. and Horvath, S. (2008) WGCNA: an R package for weighted correlation network analysis. {\em BMC bioinformatics}, 9(1), p.559.

[9] Parikshak, N.N., Gandal, M.J. and Geschwind, D.H. (2015) Systems biology and gene networks in neurodevelopmental and neurodegenerative disorders. {\em Nature Reviews Genetics}, 16(8), p.441.

[10] Campbell, P.J., Getz, G., Stuart, J.M., Korbel, J.O., and Stein, L.D. on behalf of the ICGC/TCGA Pan-Cancer Analysis of Whole Genomes Network.(2017) Pan-cancer analysis of whole genomes. bioRxiv preprint, bioRxiv:10.1101/162784.

[11] Akbarian, S., Liu, C., Knowles, J.A., Vaccarino, F.M., Farnham, P.J., Crawford, G.E., Jaffe, A.E., Pinto, D., Dracheva, S., Geschwind, D.H. and Mill, J. (2015) The PsychENCODE project. {\em Nature Neuroscience}, 18(12), p.1707.

[12] A. Anonymous. Comprehensive functional genomic resource and integrative model for the human brain.  (forthcoming publication)

[13] G. Yu, L. G. Wang, Y. Han, Q. Y. He. (2012) clusterProfiler: an R package for comparing biological themes among gene clusters. {\em OMICS} 16, 284-287.

[14] Kent, W.J., Sugnet, C.W., Furey, T.S., Roskin, K.M., Pringle, T.H., Zahler, A.M., and Haussler, D. (2002) UCSC Genome Browser:  The human genome browser at UCSC. {\em Genome Res.}, 12(6), p.996.

[15] Harrow, J. \textit{et al.}, (2012) GENCODE: The reference human genome annotation for The ENCODE Projec. {\em Genome Research}, 22, p.1760.

[16] Forbes, S.A. \textit{et al.}. (2017) COSMIC: somatic cancer genetics at high-resolution. {\em Nucleic Acids Research}, 45(D1), p.D777.

[17] Bergstra, J., Yamins, D., Cox, D.D. (2013) Hyperopt: A Python Library for Optimizing the Hyperparameters of Machine Learning Algorithms, {\em Proc. of the 12th Python in Sci. Conf. (SCIPY 2013)}, p.13.

[18] Chawla, N.V., Bowyer, K.W., Hall, L.O., Kegelmeyer, W.P. (2002) SMOTE: synthetic minority over-sampling technique, {\em Journal of Artificial Intelligence Research}, 16(1), p.321.

[19] Lemaitre, G., Nogueira, F., and Aridas, C.K. (2017) Imbalanced-learn: A Python Toolbox to Tackle the Curse of Imbalanced Datasets in Machine Learning. {\em Journal of Machine Learning Research}, p.1.

\section*{Appendix A: Datasets and network details}\label{app:1}

\normalsize

\noindent \textbf{PCAWG:} The PanCancer Analysis of Whole Genomes (PCAWG) study includes a variety of biological data types corresponding to 2,800 samples from the International Cancer Genome Consortium. To train networks for our analysis, rare variants are singled out for Skin Melanoma and Esophageal Adenocarcinoma samples. The predictive task according to which the neural networks have been trained is the prediction of somatic and germline variation co-occurrence at the gene level for 718 genes of the COSMIC census list fetched on May 08, 2018. Input data included 43 features ranging from germline variant signatures of known cancer genes alongside a set of biological features extracted from multiple data and annotation repositories, namely UCSC Genome Browser [14], Gencode v27 [15], and COSMIC [16]. Each model whose weights have been analyzed by \textit{rank projection trees} has 3 hidden layers. Number of hidden nodes (285-941), optimization algorithm (Adam or Nesterov Adam), and activation functions (Exponential or Rectified Linear Unit) for each network have been determined by automated hyperparamter optimization using the HyperOpt package [17]. Results are averaged for 5 neural networks trained on randomly stratified training datasets for each cancer type, with test performance of high precision and recall values ranging between 70\% and 83\%. To balance training datasets, we deployed the SMOTE oversampling algorithm [18] using the implementation in the imbalanced-learn Python package [19].

\noindent \textbf{PsychENCODE:}  The PsychENCODE dataset [11,12] contains bulk transcriptomics and other omics data from the prefrontal cortex of 1452 post-mortem subjects, including controls and subjects with schizophrenia, bipolar disorder, and autism. From these data, we create 10 training and testing partitions (including 640 and 70 samples respectively) of control and schizophrenia subjects, which are balanced 50-50\% for controls and cases, and balanced across train/test partitions for covariates including age, gender, ethnicity and assay.  We train neural networks with 2 hidden layers to predict a binary case/control indicator, with 100 and 400 nodes at layers 1 and 2 respectively, logistic sigmoid activations, and SGD with early stopping for training. We train separate neural networks using individual gene expression levels as inputs, and mean expression levels across modules of genes pretrained using WGCNA [7,8], pre-selecting the top 1\% and 15\% of genes/modules respectively according to the absolute Pearson correlation between the input and the binary output indicator on each training partition (resulting in 187 genes and 754 modules in each respective model). The test performance of the models averaged across partitions was 73.6\% and 66.1\% for the gene and module based models, respectively.

\end{document}